\documentclass[letterpaper]{article}
\usepackage{iccc}

\usepackage{times}
\usepackage{helvet}
\usepackage{courier}
\usepackage{amsmath}
\usepackage{graphicx}
\usepackage{url}            

\title{CLIP-CLOP: CLIP-Guided Collage and Photomontage}
\author{Piotr Mirowski, Dylan Banarse, Mateusz Malinowski, Simon Osindero, Chrisantha Fernando \\
DeepMind, London, UK \\
\tt{\{piotrmirowski, dylski, mateuszm, osindero, chrisantha\} @ deepmind.com}
}

\begin{document}

\maketitle
\begin{abstract}
\begin{quote}
The unabated mystique of large-scale neural networks, such as the CLIP dual image-and-text encoder, popularized automatically generated art. Increasingly more sophisticated generators enhanced the artworks' realism and visual appearance, and creative prompt engineering enabled stylistic expression. Guided by an artist-in-the-loop ideal, we design a gradient-based generator to produce collages.
It requires the human artist to curate libraries of image patches and to describe (with prompts) the whole image composition, with the option to  manually adjust the patches' positions during generation, thereby allowing humans to reclaim some control of the process and achieve greater creative freedom. We explore the aesthetic potentials of high-resolution collages, and provide an open-source Google Colab as an artistic tool.
\end{quote}
\end{abstract}

\maketitle

\section{Introduction}

A \emph{collage}, from the French \emph{coller}, is ``a composite image made by sticking newspaper cuttings, photographs, and other printed images onto a flat surface, often combined with paint'' \cite{art2008}. \emph{Photomontage} extends collage by manipulating and compositing photographs \cite{ades1976photomontage}. The origins of collage can be traced to the invention of paper in China, and \emph{photo-collage} was a social pastime for the Victorian upper-class \cite{nationalgalleries2019collage}, before Cubists Pablo Picasso and Georges Braque made collage into an art form \cite{art2008,greenberg1958pasted}.

In this paper, we formalize collage as a picture produced by optimizing affine spatial and color transformations of patches, where patches are manually selected, and then automatically sampled, moved around, recolored, and superimposed. We design a gradient-based \emph{Collage Generator} consisting of differentiable spatial and color transformations of patches followed by transparent or opaque superposition.

\begin{figure}[h!]
  \centering
  \includegraphics[width=0.9\columnwidth]{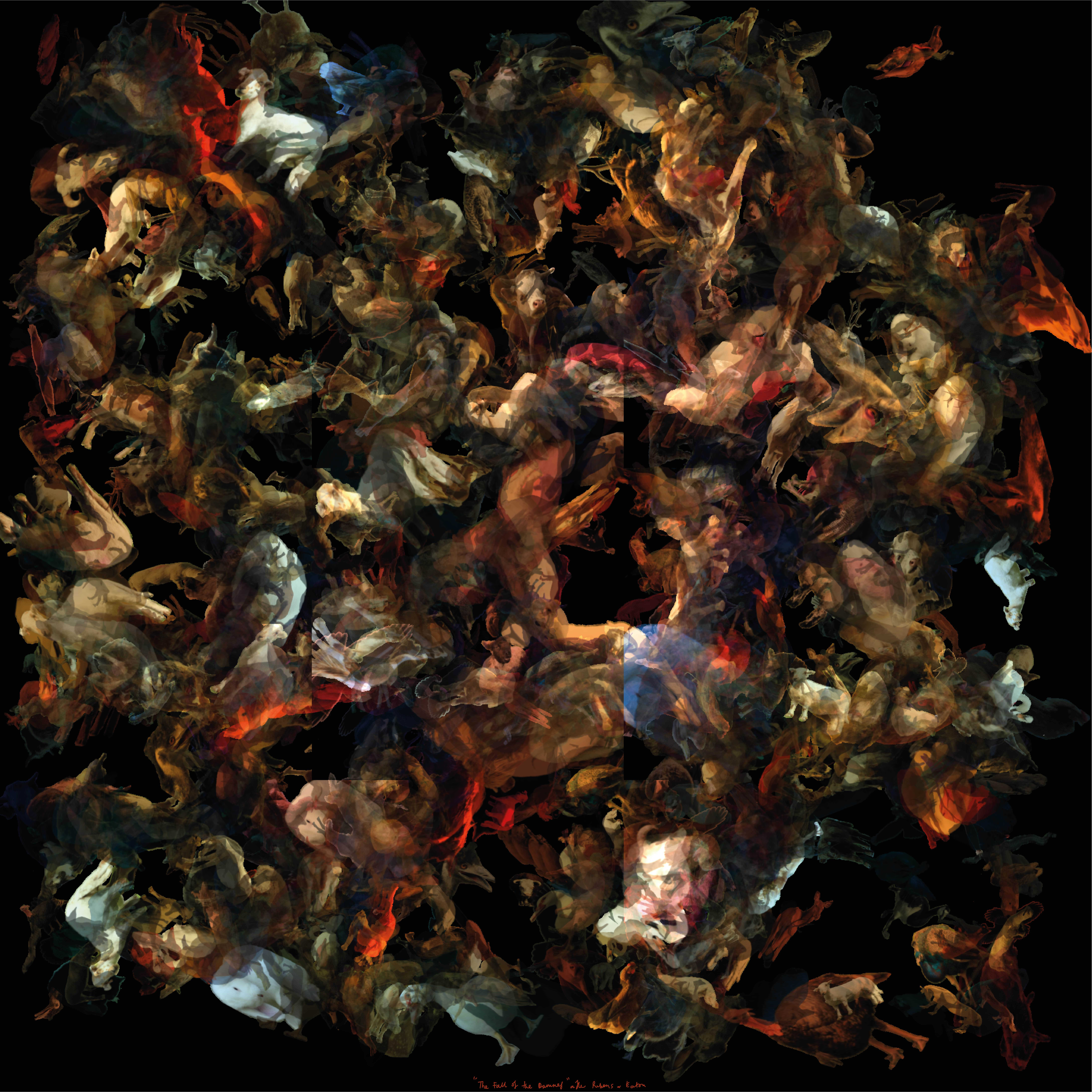}
  \caption{\emph{The Fall of the Damned after Rubens and Eaton}. High-resolution collage of image patches of animals (Fig.\ref{fig:patches}), optimized hierarchically with 3x3 overlapping CLIP critics.}
  \label{fig:damned}
\end{figure}

The \emph{Collage Generator} optimizes such transformations guided by a dual text-and-image encoder \cite{liu2021fusedream}, like the popular CLIP model from OpenAI \cite{radford2021learning}, pre-trained on large datasets of captioned images collected on the internet (hence incorporating various cultural biases). Intuitively, the dual encoder computes a score for the match between a textual \emph{prompt} and the resulting \emph{collage image}. Therefore, it acts as an AI-based \emph{Critic} assessing the ``quality'' of the artwork given its description. Large-scale dual encoders exhibit some degree of semantic compositionality, as they allow novel combinations of phrases or images, and handle such visual concepts as color, texture, shape, object relations, perspective and ``style'', to guide a generator to create remarkably convincing images. 

Computational artists like Ryan Murdock, Katherine Crowson and Mario Klingemann have investigated various neural generators, including Generative Adversarial Networks \cite{brock2018large,esser2021taming}, Diffusion models \cite{dhariwal2021diffusion}, evolution strategies on colored shapes \cite{tian2021modern} or evolution-based Neural Visual Grammars \cite{fernando2021generative}; each producing distinctive aesthetics in tandem with the CLIP critic. In Spring 2021, open-source Google Colabs allowing practitioners to combine VQGAN generators with CLIP critics \cite{crowson2022vqgan} greatly popularised the technique. More recent methods that rely on Latent Diffusion  conditioning or direct prediction of CLIP image embeddings manage to forgo the lengthy CLIP critic iterations and allow considerably faster and higher quality text-to-image generation \cite{rombach2021highresolution,ramesh2022hierarchical}.

Our system is more interpretable, as it merely optimises color and affine transformations of hand-selected patches, instead of optimizing latent variables that condition a neural pixel image generator. Since the \emph{Collage Generator} operates on collections of identifiable patches, we can let the user intervene during the optimization, manually adjusting the arrangement (shift, scale and rotation) of individual patches, for additional human-in-the-loop guidance.

Our work extends differentiable scalable vector graphics used in CLIPDraw \cite{frans2021clipdraw}, substituting strokes with patches. We experiment with various rendering methods for superimposing patches in a learnable way. We can also combine multiple critic evaluations on overlapping regions of a larger image to produce high-resolution\footnote{Open-source code and examples of high-resolution images:\\\url{https://github.com/deepmind/arnheim}} and detailed collages, allowing the artist control over the composition of the artwork. We call our system \emph{CLIP-CLOP} (loosely CLIP-guided COLlage and Photomontage) and open-source its Google Colab code.

CLIP-CLOP also builds upon extensive prior work in computational creativity. Automated collage generation \cite{krzeczkowska2010automated} in the \emph{The Painting Fool} \cite{colton2008creativity} employed keyword retrieval-based methods to make thematic news-based juxtapositions of images. Optimisation methods were used for spatial and colour transformations of image \emph{cutouts} to assemble ``Arcimboldo''-like collages that match a target image \cite{huang2011arcimboldo}. Semantic composition of patches were conceptualised as relative positions between image cutouts in \cite{breault2013soilie}, and then explored as juxtaposition, replacement and fusion of images in \cite{xiao2015vismantic} or as visual blending of emoji pictograms in \cite{cunha2018emojinating}. CLIP-CLOP combines all above aspects with differentiable transformers and a CLIP critic for textual prompt-driven image generation.

CLIP-CLOP is directly inspired by art theory. First, CLIP-CLOP arranges disparate collections of textured patches into new images, just like collage techniques enabled Cubist artists to exploit ambiguities arising from the shapes and perspectives of patches \cite{art2008}. Second, Hans Arp's \emph{Collage With Squares Arranged according to the Law of Chance} (1916-1917)\footnote{Museum of Modern Art, New York:\\ \url{https://www.moma.org/collection/works/37013}}, is a precursor to CLIP-CLOP's random initialization and optimization of patches, with optional manual adjustment, and an illustration of our human-in-the-loop approach to procedural art generation.
We believe that allowing patch choice gives the artist more control than the mere combination of prompt engineering with critic-guided generators~\cite{radford2021learning} and situates CLIP-CLOP with recent human-in-the-loop works.

\begin{figure}[h!]
\centering
\includegraphics[width=0.9\columnwidth]{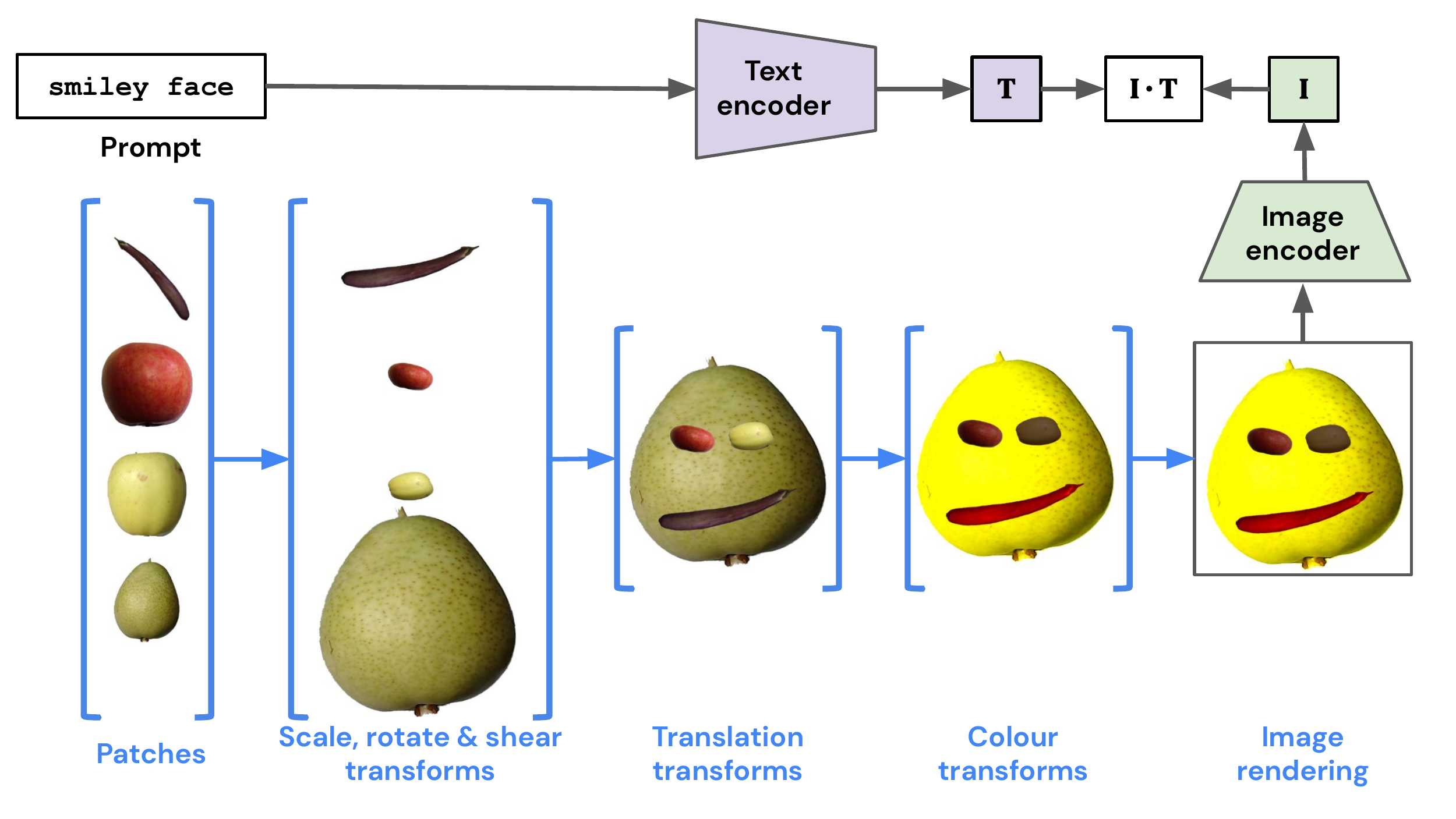}
\caption{Architecture of the generative collage algorithm.}
\label{fig:methods}
\end{figure}

\section{Algorithm}

Like in traditional \emph{collage}, the artist prepares a collection of $N$ image patches. CLIP-CLOP randomly initialises $N$ RGB color transformation vectors and  $N$ affine transformation matrices, one for each patch, to randomly color and disperse them onto a canvas. These patch transformations, as well as the patch superposition and image rendering method, constitute the \emph{Collage Generator}. A forward pass through this \emph{Collage Generator} applies transformations to each patch and then combines patches by superimposing their RGB images onto a blank canvas. The resulting image is then evaluated by the \emph{Critic} (dual image-and-text encoder) and matched to one or several user-defined textual prompts. The automated optimization loop between the parameters of the \emph{Collage Generator} and the \emph{Critic}'s evaluation is illustrated on Figure \ref{fig:methods}. An optional evolution-based optimization can be applied to the set of image patches.

\subsection{Preparation of Image Patches}
\label{patches}

We leave that crucial curation process to the artist, as it uniquely defines the artwork, and offer only basic tools for preparing image datasets.

CLIP-CLOP takes as input a list of $N$ 4-channel (RGB plus alpha) image arrays. The contour of each patch is specified by the alpha channel, which can be semi-transparent. We explored manual image segmentation using photo editing software, automated flood-filling from four corners of each patch image (when those images are adequately centered photographs of objects on a plain background) 
and computer vision-based image segmentation of photographs  over cluttered backgrounds.

\subsection{Collage Generator}
\label{generator}

The \emph{Collage Generator} is composed of color transformation, spatial affine transformation of each patch, and patch superposition, three operations that are differentiable, allowing gradient-based optimization.

\subsubsection{Color Transformation}
\label{color}

Each given image patch is assigned three color multipliers, for the red, green and blue channels; changing those coefficients with values smaller than 1 uniformly changes the patch's color. These parameters are optimized during training.

\subsubsection{Spatial Transformations}
\label{affine}

Similarly, each image patch is assigned six numbers, for X and Y translation, rotation, scale, squeeze and shear along the X axis. The two-dimensional affine transformations are expressed as $3 \times 3$ translation, rotation and scale matrices, and are applied to the image pixel 2D coordinates. The resulting affine-transformed (rotated, scaled and translated) grids of pixel coordinates are then used to interpolate the patch. Similarly, these affine transform parameters are optimized during collage generation.

\subsubsection{Differentiable Rendering}
\label{rendering}

Collage artworks typically superimpose opaque scraps of paper cuttings, assuming a partial ordering -- which scrap is on top of another. Yet in our case of differentiable rendering, a completely opaque superposition of patches compromises the learnability of the collage because we cannot propagate gradients through occluded patches.

We thus investigated two alternatives. The first one, \emph{transparency}, simply adds the RGB values of all patches. A variation of \emph{transparency}, called \emph{masked transparency}, sums RGB values of only non-masked parts of patches (i.e., where alpha channel values are strictly greater than 0) and normalizes each pixel value by the sum of masks at that position.

The second one, called \emph{opacity}, replaces the opaque, ordered superposition of patches by a differentiable approximation, consisting of a weighted sum of patches with learnable patch weights. Specifically, each patch is given an order parameter. For each patch and pixel coordinate, we compute the weight by multiplying the order of the patch by the mask at that pixel. The resulting image is a weighted sum of patches. Again, patch order parameters are optimized during collage generation.

\begin{figure}[h!]
	\centering
	\includegraphics[width=0.75\columnwidth]{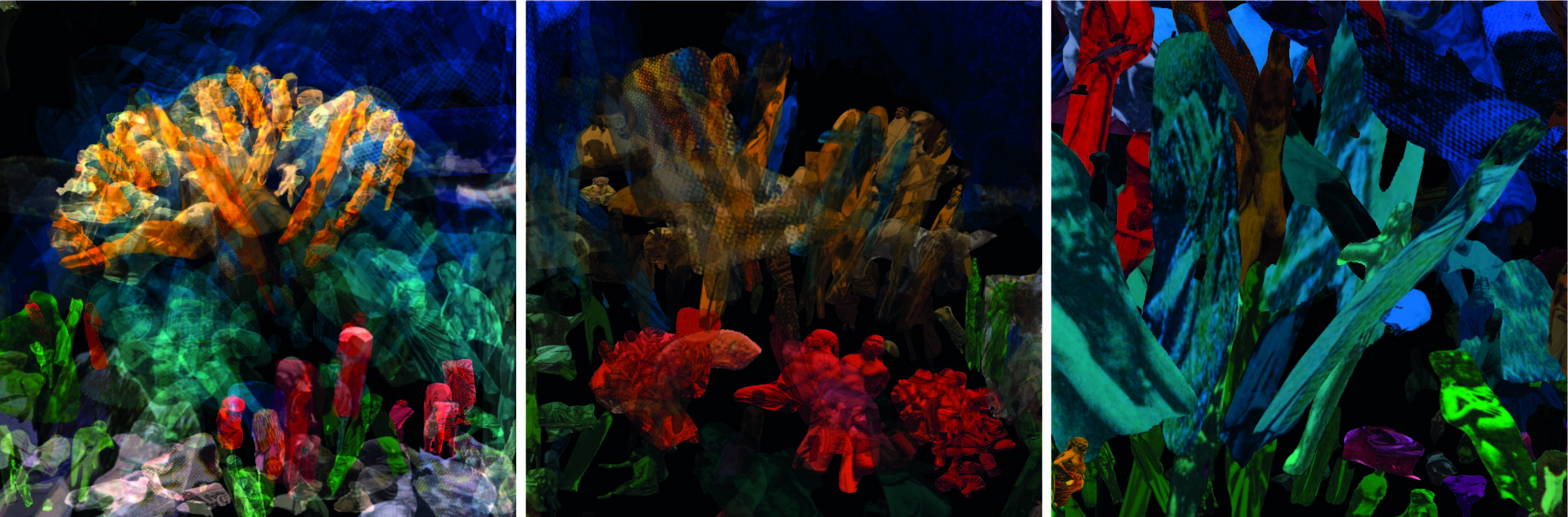}
	\caption{Rendering methods on prompt \emph{underwater coral}. Left to right: transparency, masked transparency, opacity.}
	\label{fig:rendering}
\end{figure}

Figure \ref{fig:rendering} shows the effect of various rendering methods. Note that \emph{opacity} does not always fully occlude all image patches but does more so than \emph{transparency} and \emph{masked transparency}, and that (unless the RGB range is allowed to be negative) \emph{transparency} can result in saturated (close to white) image parts (visible on left image in some of the coral tentacles). 


\subsubsection{Achieving High Resolution}

CLIP-CLOP's advantage is that it can produce collages at any resolution. During optimization, we use down-sampled patches, as CLIP is restricted to 224 x 224 images; for the final rendering, the same spatial transformations are applied to the original high resolution patches.

\subsection{Critic}
\label{critic}

By a crude analogy to an art critic, who interprets and evaluates a piece of art, we use a compatibility function -- called \emph{Critic} -- between the collage and the textual prompts. Intuitively, the higher the score given by the \emph{Critic} function, the better the fit between textual and visual inputs. At each step, \emph{Collage Generator} produces parameterized transformations of patches rendered on the canvas and generates a new collage proposal. Next, we use CLIP~\cite{radford2021learning} -- a large-scale model, trained on 400 million image-text pairs -- as \emph{Critic}, with an encoder that extracts image features from the collage and text features from the prompt. These features are matched against each other to give a compatibility score.

During training, that score is optimised by stochastic gradient ascent and backpropagated through the image to the \emph{Collage Generator} to optimize the patches' transformations parameters.

\subsubsection{Semantic Composition}
\begin{figure}[h!]
\centering
\includegraphics[width=0.66\columnwidth]{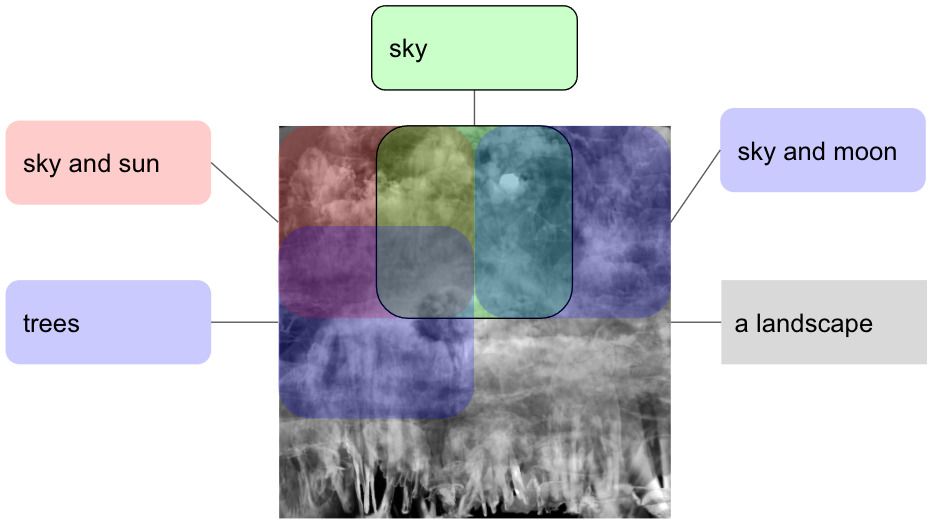}
\caption{Using multiple overlapping CLIP evaluators with different prompts allows greater control over composition and higher resolution collages.}
\label{fig:multiclip}
\end{figure}

Many generative approaches produce images with single semantics, i.e., evaluated globally with one \emph{Critic}. To achieve a higher level of compositionality, we divide the image into $3\times3$ overlapping local regions, each evaluated by a different \emph{Critic} and prompt. A tenth \emph{Critic} evaluates the whole collage globally (with reduced resolution). Figure~\ref{fig:multiclip} illustrates how one could decompose ``landscape'' using prompts: ``sky with sun'', ``sky'' and ``sky with moon'', ``trees'', etc.

Moreover, the same procedure allows to increase the resolution of the final collage. With $3 \times 3$ regions, we can produce $448 \times 448$ images instead of $224 \times 224$, typical of approaches that use CLIP. In our work, we experiment with parallel \emph{Critic} evaluations and less memory consuming but slower serial evaluations. We use either arithmetic or harmonic mean of all individual \emph{Critic} losses.

\subsubsection{Evolution of Image Patches}
\label{evolution}

Gradients enable existing patches to be manipulated but do not provide a signal for exchanging one patch for another. To support this, we optimize a population of 2 to 10 randomly initialized collages and apply a step of evolution (microbial genetic algorithm \cite{harvey2009microbial}) every 100 gradient descent steps. The scores of two random \emph{Collage Generators} are compared and the loser is overwritten with the winner, with random mutations involving swapping a random patch for another or small Gaussian noise added to affine and color transformations.

\section{Explorations}

\subsubsection{Non-Semantic Composition from Patches}
\label{exaptation}

\begin{figure}[h!]
\centering
\includegraphics[width=0.75\columnwidth]{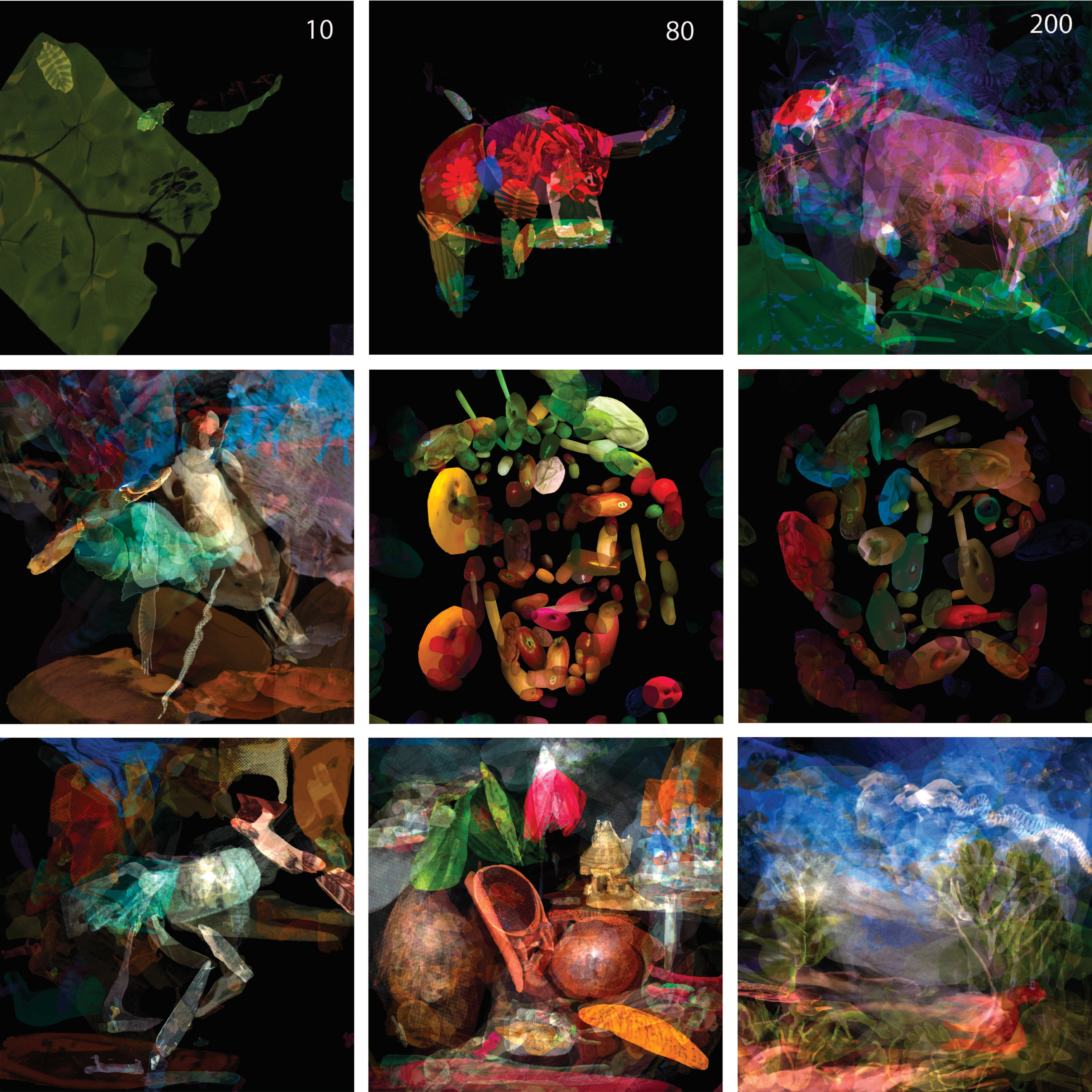}
\caption{Collages made of different numbers of tree leaves patches (bulls in the top row), as well as Degas-inspired ballet dancers made from animals, faces made of fruit, and still life or landscape made from patches of animals (see Fig. \ref{fig:patches}).}
\label{fig:bulls_ballet_faces_nature}
\end{figure}

In many human-made collages, the arrangement of patches is determined by their semantic relationship, e.g. a giant child may be depicted climbing atop a skyscraper. The meaning of each part is coherent or interestingly incongruous, and humans can easily construct such scenes. However, a harder task for humans is to compose an image (e.g. a bull or a human face\footnote{A Twitter bot regularly posts collages of human faces, generated by CLIP-CLOP and using patches of animals or human-made waste at: \url{https://twitter.com/VisPlastica/media}}) from semantically different parts (e.g. tree leaves or fruits), as illustrated on Figure \ref{fig:bulls_ballet_faces_nature}. CLIP-CLOP easily makes such discoveries and compose patches in a non-semantic way. Figure~\ref{fig:bulls_ballet_faces_nature} also shows (in the top row) that fewer patches make more abstract Picasso-inspired collages of a bull, while more patches make more realistic images.

\begin{figure}[h!]
\centering
\includegraphics[width=0.24\columnwidth]{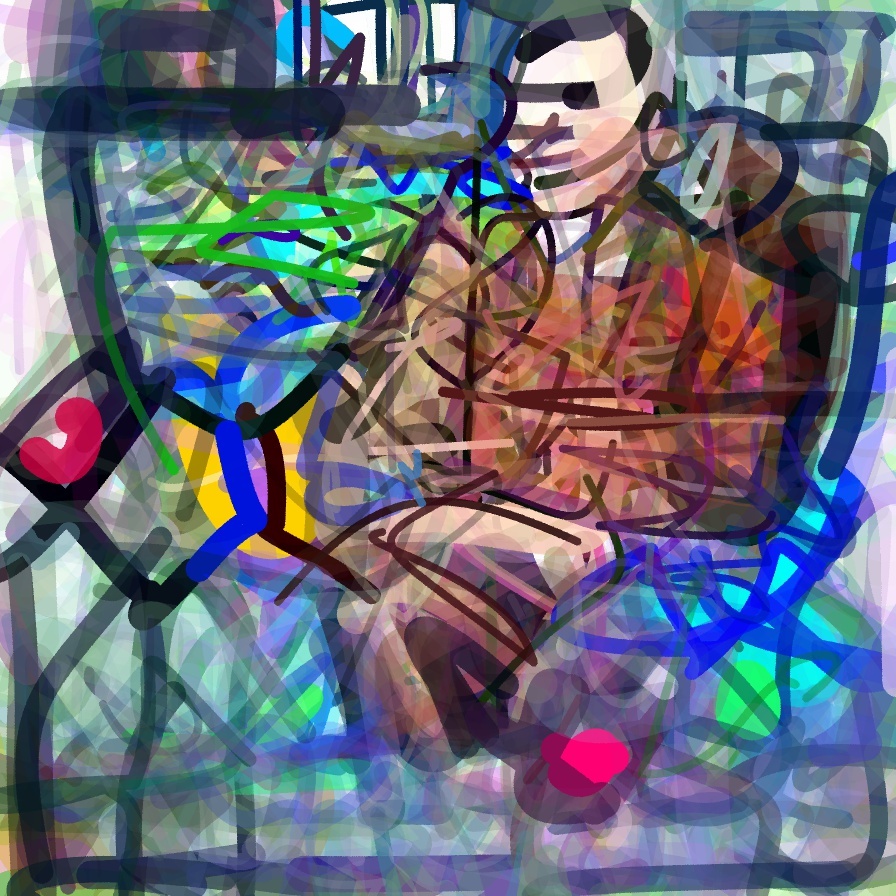}
\includegraphics[width=0.24\columnwidth]{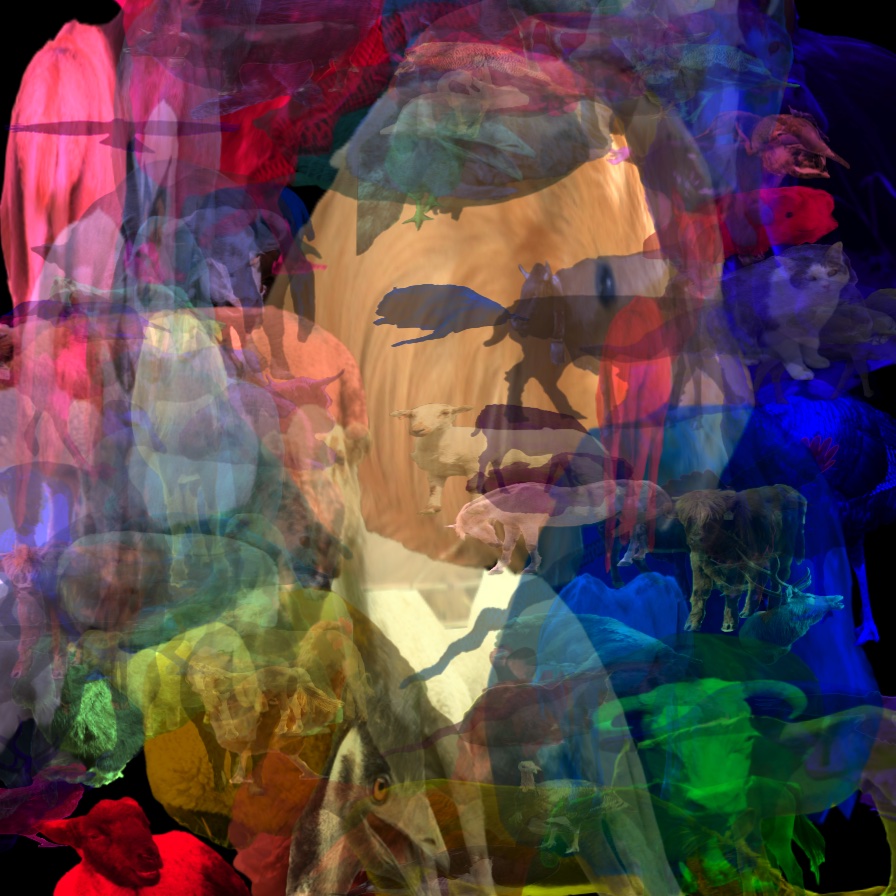}
\includegraphics[width=0.24\columnwidth]{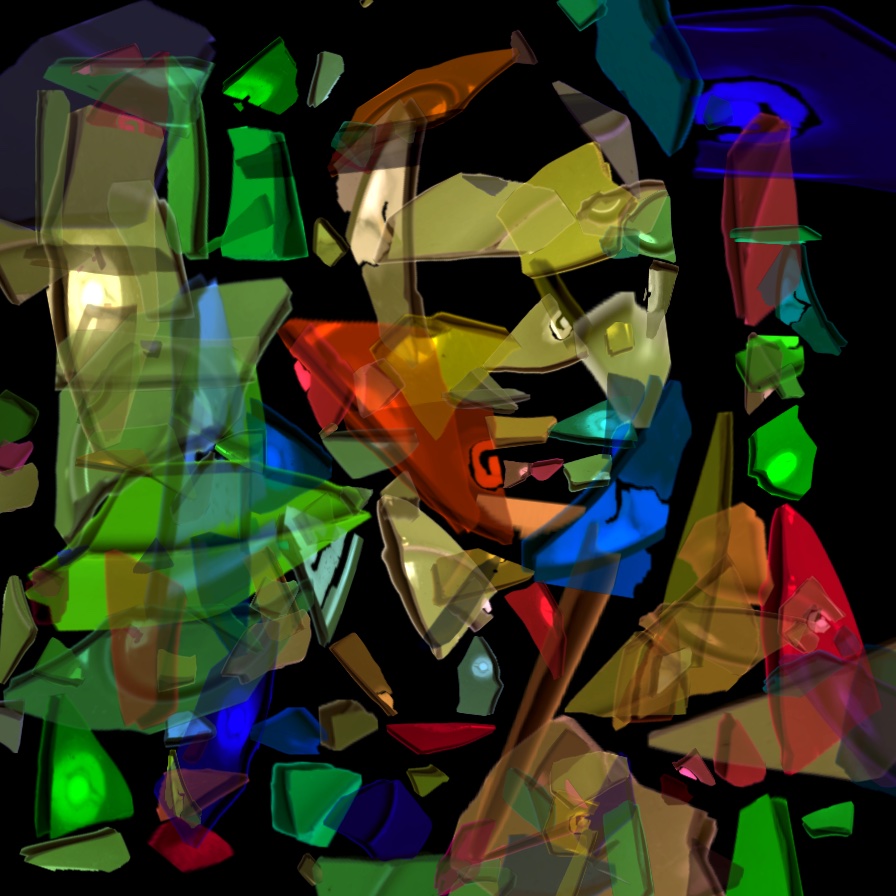}
\includegraphics[width=0.24\columnwidth]{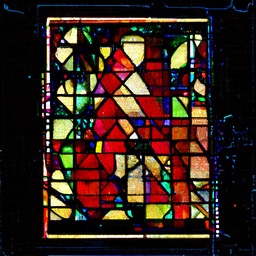}
\caption{\emph{Alan Turing in stained glass}. a) CLIPDraw with 1000 strokes, b) CLIP-CLOP with 100 animal patches or c) 200 broken plate patches, d) CLIP-guided diffusion.}
\label{fig:stainedglass}
\end{figure}

\subsubsection{Patches as Textural Image Constituents}
\label{texture}

Figure \ref{fig:stainedglass} illustrates different aesthetics that can be obtained using diverse image generators on the same prompt (\emph{Alan Turing in stained glass}). Without cherry-picking, we compared results on CLIPDraw \cite{frans2021clipdraw}, Katherine Crowson's CLIP-guided diffusion \cite{dhariwal2021diffusion} using a Colab from late 2021, and CLIP-CLOP on patches consisting of animals or fragments from a broken plate. We noticed that CLIPDraw combines many strokes to create textures, guided diffusion generates complex textures directly from pixels, while collage exploits existing shapes, shadows and textures present on individual patches to achieve the desired stained glass effect.



\subsubsection{Generative Collage as a Human-in-the-loop AI}
\label{discussion}

Popular applications such as \emph{Wombo Art}\footnote{\url{https://app.wombo.art}} that rely on state-of-the-art deep learning for image synthesis, have democratised the use of generative art systems but also removed the human user from most of the image production process, letting them only specify the prompt, and therefore focusing users' input on creative prompt engineering \cite{liu2022design}. The user has limited choice in how to visually represent concepts, cannot control the various cultural references and merely acts as a curator of the outputs \cite{chung2021human}. In a departure from commoditized art generation, we propose to give the artist full control over the image patches used for collage, making them curator of the inputs for the algorithm and collaborator with machine creativity.
We believe in art as means of human agency, requiring that ``automation in creative fields is always supported by the development of humans' creative potential'' \cite{daniele2019ai+}, and thus favour interactive systems over fully automated ones.

Human-in-the-loop systems such as \emph{collabdraw} \cite{fan2019collabdraw} or Drawing Apprentice \cite{davis2016empirically} have long been used for AI-guided sketching, and it was found that "AI Steering Tools" for musical composition that let users constrain the generative process "helped users increase their control, creative ownership, and sense of collaboration with the generative ML model" \cite{louie2020novice}.

In that spirit, we added a simple interactive human-in-the-loop correction of AI-optimized collage. We allow the artist to stop the optimization loop, manually edit one or more patches via a user interface (click on the current collage to select a patch, and adjust its position, rotation, scale, etc. using UI sliders) and then resume the optimization loop.



\section{Conclusion}

The remixability of modern media encourages sampling and remixing, hence: collage \cite{manovich2005remixing}. Collage is yet a little-explored art form for procedural visual art generation. In our work, we introduce a \emph{Collage Generator} and combine it with a popular dual image-and-text encoder like CLIP for AI-based steering. The ability to choose image primitives gives the artist an unprecedented level of control compared to previous CLIP-guided methods and helps to escape, to some extent, the straight-jacket of style imposed by pre-trained neural network generators. Current development work focuses on real-time manipulation of image patches during optimization. We resisted going in the opposite direction: automating the image primitive selection process.
We open-source\footnote{\url{https://github.com/deepmind/arnheim}} CLIP-CLOP as a creative tool for artists. 

\section{Acknowledgements}

The authors wish to thank Kitty Stacpoole, Christina Lu, Luba Eliott, Max Cant, Jordan Hoffmann, Oriol Vinyals, Ali Eslami, El Morrison, Kory Mathewson, Ian Horton, Mikołaj Bińkowski, Wojciech Stokowiec, Sina Samangooei, JB Alayrac, Yotam Doron, DeepMind colleagues and anonymous reviewers for helpful discussions and suggestions.

\section{Author Contributions}

P.M., C.F. and S.O. conceived of the presented idea. P.M., D.B., C.F. and M.M. developed the code. C.F., D.B., P.M. and M.M. generated the artworks. P.M., D.B. and C.F. collected image patches. S.O. supervised this work. All authors contributed to the final manuscript.

\section{Image Patches Used in CLIP-CLOP Collages}

\begin{figure}[h!]
\centering
\includegraphics[width=1.0\columnwidth]{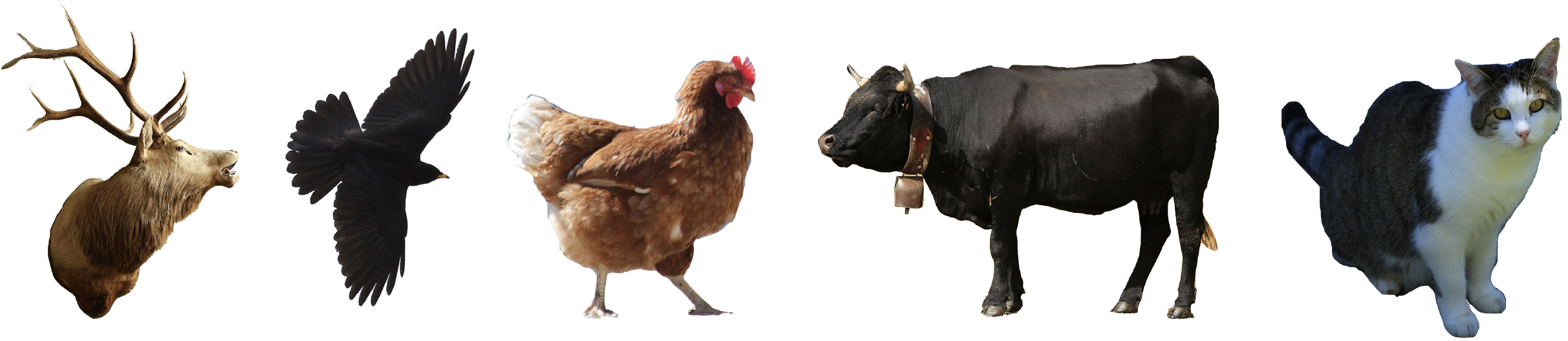}
\caption{Examples of animal patches. Photos: P. Mirowski. CLIP-CLOP is distributed with copyright-free images (public domain, CC, or photographed or drawn by the authors), see: \url{https://github.com/deepmind/arnheim}}
\label{fig:patches}
\end{figure}


\makeatletter
\def\@biblabel#1{}
\makeatother

\bibliographystyle{iccc}
\bibliography{collage}

\end{document}